%% file: main.tex
\documentclass[letterpaper]{article}
\usepackage[preprint]{aaai2027}
\usepackage[hyphens]{url}
\usepackage{graphicx}
\usepackage{natbib}
\usepackage{caption}
\usepackage{listings}
\usepackage{booktabs}
\usepackage{amsmath,amssymb}
\usepackage[hidelinks]{hyperref}

\definecolor{grouporange}{RGB}{213,94,0}
\definecolor{groupgreen}{RGB}{0,158,115}
\definecolor{refgray}{gray}{0.45}
\definecolor{royalblue}{RGB}{65,105,225}
\newcommand{\refval}[1]{\textcolor{refgray}{#1}}

\newif\ifshowrevisions
\showrevisionsfalse
\definecolor{revfps}{RGB}{0,84,200}
\definecolor{revnum}{RGB}{215,0,0}
\definecolor{revtext}{RGB}{235,120,0}
\newcommand{\revfps}[1]{\ifshowrevisions\textcolor{revfps}{#1}\else#1\fi}
\newcommand{\revnum}[1]{\ifshowrevisions\textcolor{revnum}{#1}\else#1\fi}
\newcommand{\revtext}[1]{\ifshowrevisions\textcolor{revtext}{#1}\else#1\fi}
\newcommand{\revfpson}{\ifshowrevisions\color{revfps}\fi}
\makeatletter
\def\rev@bibkeys{,BasicVSR,CAP_VSTNet,CCPL,Coherent_Video_Style,D_LUT,Deep_Preset,Linear_Transformations,Lumivideo,MCCNet,MROPM,ModFlows,NLUT,Neural_Preset,PhotoWCT,PhotoWCT2,Restoration,Restoration2,Restoration3,SA_LUT,Temporal_Color_Transfer,Ultrafast_Photorealistic,Universal_Style_Transfer,Video_color_grading,bilateral_grids,cdc,cogvideox,colorization,colorization2,colorization3_BiSTNet,condensedmovies,deqa,dinov2,fluxkontext,ldc,musiq,qalign,qwen3vl,ssim,transnetv2,wanvace,}
\newcommand{\rev@bibcolor}[1]{\ifshowrevisions\normalcolor
  \@expandtwoargs\in@{,#1,}{\rev@bibkeys}\ifin@\color{revtext}\fi\fi}
\AtBeginDocument{\let\rev@orig@bibitem\bibitem
  \renewcommand{\bibitem}{\@ifnextchar[\rev@lbibitem\rev@bibitem}}
\def\rev@lbibitem[#1]#2{\rev@bibcolor{#2}\rev@orig@bibitem[{#1}]{#2}}
\def\rev@bibitem#1{\rev@bibcolor{#1}\rev@orig@bibitem{#1}}
\makeatother

\title{From Legacy to Contemporary: Scalable Color Modernization for Long Videos}
\author{\revtext{
    Dongyun Kim\thanks{Equal contribution.},
    Jiwoo Choi\footnotemark[1],
    Jaeseong Lee,
    Seon Joo Kim}
}
\affiliations{
    \revtext{Yonsei University}\\
    \revtext{\texttt{kdyun01@yonsei.ac.kr jiw55@yonsei.ac.kr}}
}

\begin{document}

\maketitle
\begin{abstract}
Bringing legacy video back to contemporary screens does not necessarily make it look contemporary. Earlier imaging pipelines often leave color casts, muted chroma, weak contrast, and tonal imbalance, while correcting these characteristics still relies heavily on expert colorists and scales poorly to long, multi-shot videos. We formulate long video color modernization, a task that updates the chromatic and tonal appearance of legacy color footage while preserving scene content and coherence across frames and shots. We introduce a scalable, training-free framework that jointly modernizes grids of temporally adjacent keyframes using a pretrained text-guided image editing model, producing coherent contemporary references without paired supervision. Compact shot-specific color transforms then propagate these adjustments throughout each shot, preserving spatial structure and temporal stability. We conduct a multidimensional evaluation on diverse legacy videos spanning five decades. We combine a human evaluation of perceived modernness with quantitative measures of content fidelity, temporal stability, perceptual quality, and efficiency, and introduce Source-Weighted Inter-shot Grade Consistency (SW-IGC) to assess grade coherence across shots. On our evaluation set, our framework achieves the highest perceived modernness while remaining competitive across complementary measures, opening a path toward the systematic study of long video color modernization. Code is available at {\href{https://github.com/jiwoo-mood/Video_Color_Modernization.git}{\textcolor{blue!55!cyan!80}{\textit{\nolinkurl{https://github.com/jiwoo-mood/Video_Color_Modernization.git}}}}}
\end{abstract}

\section{Introduction}

Legacy videos continue to return to contemporary screens through restoration, redistribution, and archival release. Yet their return alone does not ensure that they meet the visual expectations of today's viewers. The issue is often not what they depict, but the color appearance inherited from earlier imaging pipelines. Limited capture devices, analog or early digital processing, and broadcast transfer can leave persistent color casts, desaturation, reduced contrast, and tonal imbalance. These color and tonal degradations make otherwise valuable footage feel visually dated, motivating color modernization methods that preserve scene content and structure.

Growing interest in AI-assisted remastering, automatic color correction, and reference-based look matching reflects the practical demand for such modernization. However, high-quality results still typically rely on expert colorists to adjust exposure, contrast, and chroma while maintaining consistency across frames and shots. Since this process depends on aesthetic judgment and technical expertise, it scales poorly to long videos with diverse scenes and lighting. These limitations call for automated color modernization that can scale without compromising video coherence.

We formalize this problem as \textbf{long video color modernization}: adjusting the color and tone of legacy footage toward a broadly contemporary appearance while preserving scene content and structure. Unlike restoration~\citep{BasicVSR,Restoration,Restoration2,Restoration3}, which aims to recover degraded visual signals, and colorization~\citep{colorization,colorization2,colorization3_BiSTNet,colorization4}, which infers missing chroma for grayscale footage, our goal is to modernize footage that already contains color yet appears visually dated.

This formulation raises two technical challenges and one evaluation challenge. First, image-level modernization is underdetermined. Modern color is not a unique target: multiple color grades may be perceived as contemporary, and paired old-to-modern supervision is generally unavailable. Second, long videos require color transformations that are both shot-adaptive and video-coherent. Frame-dependent transformations can fluctuate with content and introduce flicker, whereas a single global transformation can fail to account for shot-specific lighting and color statistics. Effective modernization must therefore remain stable within each shot while avoiding arbitrary grade drift across visually related shots. Third, to our knowledge, evaluating long video color modernization remains underexplored. Existing assessments largely focus on individual images~\citep{MROPM} or short clips~\citep{Coherent_Video_Style,MCCNet,CAP_VSTNet,NLUT,Temporal_Color_Transfer}, leaving the broader requirements of long multi-shot videos less systematically examined. A meaningful evaluation therefore requires a multidimensional assessment encompassing perceived modernness, content preservation, temporal stability, cross-shot grade coherence, perceptual quality, and efficiency.

To address these technical challenges, we propose a scalable framework that combines keyframe modernization with shot-level propagation. We first divide the input video into shots and select one representative keyframe per shot. Temporally adjacent keyframes are then packed into a grid and jointly modernized using a pretrained text-guided image editing model~\revtext{\citep{fluxkontext}}. The model’s learned generative prior provides plausible contemporary references without paired supervision, while joint processing promotes coherent grading across neighboring shots. Finally, we fit a compact, shot-specific transform to each source–modernized keyframe pair and propagate its chromatic and tonal adjustments throughout the corresponding shot. Per-shot estimation accommodates variations in lighting and color statistics, while applying the same pixelwise transform throughout each shot maintains stable grading without altering spatial structure. By applying generative editing only to representative keyframes, the framework scales efficiently to long multi-shot videos.

To systematically assess long video color modernization, we curate a diverse evaluation set of multi-shot legacy videos from the Condensed Movies \revtext{Dataset~\citep{condensedmovies}}, spanning five decades of footage. We evaluate perceived modernness through a user study and complement it with quantitative measures of content preservation, temporal stability, perceptual quality, and efficiency. We further introduce Source-Weighted Inter-shot Grade Consistency (SW-IGC), which evaluates inter-shot grade coherence by weighting shot-pair comparisons according to source color similarity. Among the eight evaluated versions, our framework achieves the best mean rank for perceived modernness, whereas several baselines do not surpass the unedited original. It also remains competitive across the complementary quantitative criteria.

Our contributions are threefold: \textbf{(i)} we formulate long video color modernization as a distinct task for updating the chromatic and tonal appearance of legacy color footage; \textbf{(ii)} we propose a scalable framework that jointly modernizes temporally adjacent keyframes and propagates their adjustments through compact, shot-specific color transforms; and \textbf{(iii)} we provide a multidimensional evaluation of this task, combining human judgments of perceived modernness with complementary quantitative measures and introducing SW-IGC for inter-shot grade coherence.

\input{figures/main/pipeline_include}

\section{Related Work}

\subsection{Reference-based Image Color Style Transfer}
Modernizing the color appearance of legacy footage is closely related to color grading, in which colorists adjust color and tone to achieve a desired visual look rather than recover a single objectively correct appearance. Because such an appearance is subjective and lacks a unique ground truth, it can be specified through a reference image and approached through color style transfer.

Previous methods transfer color from source--reference image pairs either by aligning feature statistics or by predicting color operators. Feature-statistics-based methods, such as WCT variants~\citep{Universal_Style_Transfer,PhotoWCT, Ultrafast_Photorealistic,PhotoWCT2, CAP_VSTNet}, whiten source features and recolor them using the statistics of the reference features. In contrast, color operators are commonly represented as locally affine bilateral grids~\citep{bilateral_grids} or parametric color mappings~\citep{Deep_Preset,Neural_Preset,ModFlows,D_LUT,SA_LUT}. Since these methods alter image appearance without explicitly modifying spatial structure, they generally preserve image layout and can be applied at high resolution.

Our scalable framework supports various color-propagation methods, but prioritizes deterministic color mapping over inference-time style transfer for efficient long-video processing. Our evaluation shows that this \revtext{training-free} approach provides sufficient visual quality.

\subsection{Video Color Grading}

Extending reference-based color style transfer to video requires keeping the appearance consistent across frames. Existing studies~\citep{Linear_Transformations,MCCNet,CCPL,CAP_VSTNet, Temporal_Color_Transfer} estimate a transformation operator for each frame or map \revtext{an} operator to the video without any temporal component such as optical flow or a temporal loss. These methods mainly address a single shot or a short clip, and do not explicitly consider coherent grading across the many shots of a long-form video.

Even studies that explicitly target video color grading~\citep{Video_color_grading,Lumivideo} \revtext{treat} a sequence as one continuous clip, and do not address the color inconsistency that arises within a shot or between shots.
\citet{Video_color_grading} generate a LUT learned from a source--reference pair, but apply that single LUT to the entire video, irrespective of shot boundaries. \citet{Lumivideo} use LLM-based RAG and a tree-of-thoughts procedure to determine the parameters of a single 3D LUT, which is likewise applied uniformly across the whole log video.
Recent video generation models offer flexible end-to-end manipulation~\citep{wanvace,cogvideox}, but are developed primarily for synthesizing or editing short, motion-rich clips rather than preserving the source content exactly. Their generative formulation can introduce unintended content changes and temporal inconsistencies, while the high cost of dense generation limits their applicability to long-form video modernization.

\section{Method}

\subsection{Framework Overview}
\label{sec:overview}

Given a long-form legacy color video $V = (I_t)_{t=1}^{T}$, our goal is to produce a modernized video $\widetilde{V} = (\widetilde{I}_t)_{t=1}^{T}$ with a contemporary chromatic and tonal appearance while preserving the original content and maintaining temporal consistency. As illustrated in Fig.~\ref{fig:framework}, our framework comprises three stages: keyframe extraction, image-level modernization, and color propagation. First, we partition the input video into shots and select a representative keyframe from each shot (Sec.~\ref{sec:method_1}). We then pack temporally adjacent keyframes and modernize them jointly using a pretrained text-guided image editing model to promote inter-shot consistency (Sec.~\ref{sec:method_2}). Finally, we estimate a shot-specific color mapping from each original--modernized keyframe pair and apply it throughout the corresponding shot (Sec.~\ref{sec:method_3}).

\subsection{Keyframe Extraction}
\label{sec:method_1}

To support efficient shot-level color propagation, \revtext{we} partition each video into shots and select one representative keyframe per shot through boundary detection, segment merging, and keyframe selection.

\paragraph{\protect\revtext{Shot detection \& merging.}}

We detect candidate boundaries using TransNet~V2~\revtext{\citep{transnetv2}} with a recall-oriented low threshold, preventing a single color mapping from spanning distinct shots. To correct false splits caused by flashes or rapid motion, we greedily merge adjacent segments when the DINOv2~\citep{dinov2} cosine similarity between the frames immediately surrounding their boundary exceeds a threshold. This semantic criterion helps distinguish transient within-shot appearance changes from true boundaries between distinct shots.

\paragraph{Keyframe selection.}

For each shot, we sample candidate frames and use a vision--language model~\citep{qwen3vl} to select the clearest and most representative one, since visual clarity and shot representativeness are difficult to capture with a single heuristic. This keyframe serves both as the modernization input and as the reference for shot-level color propagation.

\subsection{Image-Level Modernization}
\label{sec:method_2}

Image modernization is inherently ill-posed: there is no single ground-truth modernized result, and paired legacy and modern versions of the same visual content are generally unavailable. We therefore do not learn a direct old-to-modern mapping from paired supervision. Instead, we employ a pretrained text-guided image editing model to modernize the selected keyframes. Specifically, we use FLUX.1-Kontext~\revtext{\citep{fluxkontext}} with a fixed instruction that requests natural, contemporary color and tonal adjustments while preserving the original visual content and composition.

Processing each keyframe independently may produce inconsistent color rendering across neighboring shots. To improve inter-shot consistency, we adopt a three-step joint modernization procedure. First, during \textbf{keyframe packing}, we pack up to four temporally adjacent keyframes into a $2 \times 2$ grid. We then perform \textbf{joint keyframe modernization} by editing the packed grid in a single model invocation. By exposing the model to neighboring shots simultaneously, this joint processing encourages consistent chromatic and tonal adjustments across them. Finally, during \textbf{keyframe unpacking}, we split the edited grid into individual modernized keyframes. Each modernized keyframe is subsequently paired with its original counterpart to estimate the color mapping for the corresponding shot.

\subsection{Color Propagation}
\label{sec:method_3}

This stage consists of two components. We first estimate a shot-specific color transform from each legacy--modernized keyframe pair, and then apply it consistently to all frames in the corresponding shot with highlight roll-off.

\paragraph{Shot-specific color mapping.}

For each shot $s$, we compute a global affine mapping that transports the RGB color distribution of its selected legacy keyframe to that of its modernized counterpart. Since the modernization stage targets primarily chromatic and tonal adjustments while preserving scene content, a single global mapping provides a compact approximation that can be applied pixelwise without altering spatial structure.

Let $(\boldsymbol{\mu}_s,\boldsymbol{\Sigma}_s)$ and $(\boldsymbol{\mu}_s^{\mathrm{m}},\boldsymbol{\Sigma}_s^{\mathrm{m}})$ denote the RGB means and covariance matrices of the legacy and modernized keyframes, respectively. We compute the shot-specific transform using the Monge--Kantorovitch linear (MKL) color mapping~\citep{Monge_Kantorovitch}. Given an RGB color vector $\mathbf{x}\in\mathbb{R}^{3}$,
\begin{equation}
\label{eq:mk_mapping}
\begin{aligned}
\mathcal{M}_s(\mathbf{x})
&=
\mathbf{T}_s
\bigl(\mathbf{x}-\boldsymbol{\mu}_s\bigr)
+
\boldsymbol{\mu}_s^{\mathrm{m}},\\
\mathbf{T}_s
&=
\boldsymbol{\Sigma}_s^{-1/2}
\left(
\boldsymbol{\Sigma}_s^{1/2}
\boldsymbol{\Sigma}_s^{\mathrm{m}}
\boldsymbol{\Sigma}_s^{1/2}
\right)^{1/2}
\boldsymbol{\Sigma}_s^{-1/2}.
\end{aligned}
\end{equation}
Here, $(\cdot)^{1/2}$ denotes the principal matrix square root. Under Gaussian approximations of the legacy and modernized color distributions, $\mathcal{M}_s$ matches the target mean and covariance while minimizing the expected squared transport cost. Unlike independent per-channel mappings, the full $3\times3$ matrix $\mathbf{T}_s$ captures cross-channel correlations and enables coupled adjustments across RGB channels.

\paragraph{Shot-level color propagation.}

The affine mapping can produce RGB values outside the valid range, causing hard clipping, particularly in bright regions. We therefore apply a luma-based highlight roll-off that smoothly compresses luma values above a fixed knee using a $\tanh$-based curve and uniformly rescales the RGB channels to preserve hue. For every frame $I_t$ in shot $s$, we apply $\mathcal{M}_s$ pixelwise, followed by the highlight roll-off, to obtain $\widetilde{I}_t$. Using the same mapping and roll-off throughout the shot yields a temporally consistent color transformation across its frames.

\section{Experiments}

We evaluate whether our framework produces perceptually convincing modern color appearances while preserving source fidelity and temporal consistency. We assess modern color preference through a user study, complement it with quantitative and qualitative analyses, and examine key design choices across the framework through ablation studies.

\subsection{Experimental Setup}

\paragraph{Datasets.}

We evaluate on a curated subset of the Condensed Movies Dataset~\revtext{\citep{condensedmovies}}, a large-scale collection of key scenes from thousands of films. To capture diverse color characteristics across eras, we sample 20 videos per decade from the 1960s through the 2000s, yielding 100 videos, 3.9 hours of footage, and 3,390 shots. Each video is 2--4 minutes long; these short excerpts increase content diversity, while our framework is not limited to such durations.

\paragraph{Baselines.}

We consider two families of baselines, categorized by which stages of our framework they replace.

\noindent\textit{\textbf{End-to-end video editing.}}
We compare against two end-to-end video editors, CogVideoX-5B~\citep{cogvideox} and Wan2.1-VACE-1.3B~\citep{wanvace}, which directly modernize the input video without explicit keyframe processing or color propagation. This comparison evaluates whether dense video editing can preserve source content and temporal coherence over long videos.

\noindent\textit{\textbf{Modular color propagation.}}
The second family replaces only the propagation stage of our framework. We retain the same shot boundaries and modernized reference keyframes, and use CAP-VSTNet~\citep{CAP_VSTNet}, Neural Preset~\citep{Neural_Preset}, Video Color Grading \revtext{(VCG)}~\citep{Video_color_grading}, and SA-LUT~\citep{SA_LUT} to transfer the reference color grade to all frames within each shot. This controlled comparison isolates the effect of the propagation mechanism.

\paragraph{\protect\revtext{Evaluation metrics.}}

Video modernization is inherently a one-to-many problem, as multiple color grades can provide valid modern renditions of the same source video, precluding evaluation against a single ground-truth grade. We therefore assess perceived modern color appearance through human judgments as the primary evaluation objective, complemented by measures of structural preservation, intra-shot temporal color consistency, perceptual and aesthetic quality, inter-shot grade consistency, and computational efficiency.

\noindent\textit{\textbf{Modern color preference.}}
We conduct a blinded user study with 65 participants recruited \revtext{via Prolific} to evaluate perceived modern color appearance. Each participant evaluates the same 15 video sets in randomized order. For each video, participants simultaneously view eight anonymized versions in a grid—the output of our method, the source video, and six baseline outputs—with their positions randomized. Including the source video allows us to assess whether the edited outputs are perceived as more contemporary than the original, beyond their relative preference among competing methods. Participants rank all eight versions from 1 (best) to 8 (worst) \revtext{by modern color appearance (judged on color alone) and temporal consistency of colors and content}. The 975 rankings are summarized by \revtext{Mean Rank, Rank-1, Top-3, and the full rank distribution}.

\noindent\textit{\textbf{Structural and temporal consistency.}}
We evaluate structural preservation using Edge-SSIM~\citep{ssim,ldc} and intra-shot temporal color consistency using CDC~\citep{cdc}, which measures temporal variation in framewise color distributions.

\noindent\textit{\textbf{Perceptual and aesthetic quality.}}
We assess perceptual quality using DeQA-Score~\citep{deqa} \revnum{and the quality score of Q-Align~\citep{qalign} (Q-Align$_\mathrm{Q}$), and aesthetic quality using its aesthetic score (Q-Align$_\mathrm{A}$)}.

\noindent\textit{\textbf{Efficiency.}}
\revfps{We measure the propagation throughput in frames per second.}

\noindent\textit{\textbf{Inter-shot grade consistency.}}
We introduce Source-Weighted Inter-shot Grade Consistency (SW-IGC) to measure the consistency of shot-level grading transformations while weighting shot pairs by source color similarity. For each shot $i$, we fit an affine mapping $\mathcal{F}_i(\mathbf{x})=\mathbf{A}_i\mathbf{x}+\mathbf{b}_i$ in linear RGB using aligned source--output pixel pairs. We then apply each fitted mapping to a fixed set of $K{=}24$ ColorChecker reference colors, enabling a content-independent comparison of the resulting color transformations. We define the discrepancy between the fitted mappings for shots $i$ and $j$ as
\begin{equation}
    d_{ij}
    =
    \frac{1}{K}
    \sum_{k=1}^{K}
    \Delta E_{00}\!\left(
    \operatorname{Lab}\!\left(\mathcal{F}_i(\mathbf{c}_k)\right),
    \operatorname{Lab}\!\left(\mathcal{F}_j(\mathbf{c}_k)\right)
    \right),
\end{equation}
where $\mathbf{c}_k$ denotes the $k$-th ColorChecker reference color. Let $\mathbf{m}_i^{\mathrm{src}}$ be the channel-wise median CIELAB color of the sampled source pixels in shot $i$. The source similarity weight is defined as
\begin{equation}
    w_{ij}
    =
    \exp\!\left(
    -\frac{
    \Delta E_{00}\!\left(
    \mathbf{m}_i^{\mathrm{src}},
    \mathbf{m}_j^{\mathrm{src}}
    \right)^2
    }{
    2\sigma^2
    }
    \right).
\end{equation}
We set $\sigma$ to the median pairwise source color distance within each video, adapting the weighting scale to the source color differences across its shots. SW-IGC is then computed as
\begin{equation}
    \mathrm{SW\text{-}IGC}
    =
    \frac{
    \sum_{i<j} w_{ij}d_{ij}
    }{
    \sum_{i<j} w_{ij}
    }.
\end{equation}
Lower SW-IGC indicates greater consistency among grading transformations for shots with similar source colors.

\subsection{Comparisons}

\paragraph{\protect\revtext{Modern color preference.}}

\input{figures/main/user_study_include}

Fig.~\ref{fig:user_study_ranks} shows a clear overall preference for our method. Ours achieves the lowest mean rank of 2.31, which is 0.71 lower than that of CAP-VSTNet, the strongest baseline, and 1.31 lower than that of the source video. It receives the top rank in 37.9\% of rankings, more than twice the 17.6\% achieved by CAP-VSTNet, and places within the top three in 81.7\%, compared with 66.3\% for CAP-VSTNet. The two end-to-end editors receive the worst mean ranks, highlighting source preservation as essential to successful modernization. Neural Preset and SA-LUT rank below the source despite using the same modernized keyframes, showing that keyframe quality alone does not ensure effective video-level results.

\paragraph{\protect\revtext{Complementary quantitative results.}}

\input{tables/main/quant}

Tab.~\ref{tab:quant} complements the human evaluation with quantitative measures of structural fidelity, temporal consistency, visual quality, inter-shot grade consistency, and computational efficiency. The end-to-end video editing methods substantially alter the source content, yielding markedly lower Edge-SSIM scores of 0.439--0.535, compared with 0.745--0.886 for the modular methods. They also obtain lower perceptual and aesthetic quality scores and \revfps{generate only about 0.6 frames per second}, indicating the difficulty of preserving source fidelity and efficiency through dense video regeneration. Within the modular family, our method remains competitive in structural fidelity and visual quality. It achieves the best SW-IGC of 5.90, indicating more consistent grading transformations across shots with similar source colors. \revfps{Its training-free propagation runs at 816 FPS, on par with Neural Preset, about $10\times$ faster than CAP-VSTNet, and over $35\times$ faster than Video Color Grading and SA-LUT.} Overall, our method combines competitive source fidelity with improved inter-shot grade consistency and high computational efficiency, enabling practical modernization of long-form videos.

\paragraph{\protect\revtext{Qualitative results.}}

\input{figures/main/qual_include}

Fig.~\ref{fig:qual} shows qualitative comparisons among the evaluated methods. End-to-end video editing methods often reduce source fidelity. \revtext{CogVideoX-5B} alters facial identity and object appearance, while \revtext{Wan2.1-VACE-1.3B} removes fine details and produces blurry, desaturated outputs. Modular color-propagation methods better preserve structure but introduce distinct grading biases: Neural Preset lowers contrast, Video Color Grading oversaturates colors, SA-LUT produces dark outputs with a red cast, and CAP-VSTNet applies inconsistent grading across shots. In contrast, our method preserves source fidelity, reduces dominant color casts, and maintains natural skin tones with balanced contrast and saturation across shots.

\subsection{Ablation Studies}

\paragraph{\protect\revtext{Inter-shot consistency in keyframe modernization.}}

\input{tables/main/ablation_1}
\input{figures/main/ablation_1_include}

We ablate design choices for improving inter-shot color consistency in keyframe modernization. \emph{Independent} modernizes each keyframe in a separate model invocation, while \emph{Independent + post-hoc smoothing} temporally smooths the estimated shot-specific color mappings before propagation. For joint keyframe modernization, we arrange up to four keyframes into a $2\times2$ grid and compare two grouping strategies. \emph{Packing (clustered)} groups keyframes by source color similarity, whereas \emph{Packing (consecutive)} groups temporally adjacent keyframes.

As shown in Tab.~\ref{tab:ablation_1}\revtext{, independent} modernization produces the highest SW-IGC, indicating large variation in inter-shot color mappings. Post-hoc smoothing reduces SW-IGC most effectively, but retains full inference cost and cannot coordinate keyframe appearances during generation. Fig.~\ref{fig:ablation_1} visualizes the resulting inconsistencies and smoothing artifacts.
The two packing strategies achieve comparable SW-IGC, indicating little benefit from explicit source color clustering for overall inter-shot consistency. Moreover, clustering requires comparing keyframes across the entire video and introduces a separate grouping stage and associated hyperparameters, whereas consecutive packing follows the temporal order directly. We therefore adopt \emph{Packing (consecutive)} as the simpler and more efficient design, without sacrificing inter-shot consistency.

\paragraph{\protect\revtext{Operator design in color propagation.}}

\input{tables/main/ablation_2}
\input{figures/main/ablation_2_include}

We examine the design of the shot-specific color propagation operator. \emph{Diagonal} independently matches the mean and standard deviation of each channel, \emph{MKL} matches the full covariance to model cross-channel correlations, and \emph{Histogram} performs independent per-channel CDF matching. We further evaluate \emph{MKL + roll-off}, which augments the MKL mapping with a luma-based highlight roll-off before final clipping to the valid RGB range. For the quantitative evaluation, each mapping is applied to the source keyframe used for fitting. We compare the resulting keyframe with its modernized counterpart using the sliced-Wasserstein distance (SWD) in CIELAB for reference fidelity, and with the original source keyframe using Edge-SSIM for structural preservation.

Tab.\ref{tab:ablation_2} shows a trade-off between reference fidelity and structural preservation. \emph{Diagonal} retains high Edge-SSIM but yields the largest SWD, indicating that independent channel-wise statistics have limited expressiveness for reproducing the target color distribution. In contrast, \emph{Histogram} achieves the smallest SWD by independently matching the CDF of each source channel to that of the corresponding channel in the modernized keyframe, but yields the lowest Edge-SSIM. Fig.~\ref{fig:ablation_2}(A) qualitatively illustrates this trade-off. By modeling cross-channel correlations, \emph{MKL} reduces SWD by $9.6\%$ relative to \emph{Diagonal} while maintaining nearly identical Edge-SSIM, providing a more favorable balance between the two criteria. \emph{MKL + roll-off} performs comparably to \emph{MKL} in the fitting-keyframe evaluation while reducing clipping-induced highlight saturation on propagated frames, as shown in Fig.~\ref{fig:ablation_2}(B). We therefore adopt it as our final propagation operator.

\section{Conclusion}
In this paper, we propose a scalable framework for \revtext{long video} color modernization. Our framework jointly modernizes packed keyframes from neighboring shots using the generative prior of a pretrained \revtext{text-guided image editing} model. It then propagates the modernized colors throughout each shot using a deterministic, shot-specific color mapping. Extensive experiments show that our results are perceived as having the most contemporary color appearance. Our framework also achieves the best inter-shot color consistency among the compared methods \revfps{using lightweight, training-free color propagation}, while remaining competitive on the other quantitative metrics. These results demonstrate a practical approach to modernizing long-form video archives without compromising their original content or coherence.

\bibliography{references}

% Appendix
\twocolumn[{%
    \centering
    {\LARGE\bf From Legacy to Contemporary: Scalable Color Modernization for Long Videos\par}
    \vskip 0.1in
    {\Large\bf \revtext{Appendix}\par}
    \vskip 1.5em
}]
\makeatletter\global\@topnum\z@\makeatother % as \maketitle does: no top float in the first column
\appendix
\input{supp}

\end{document}

%% file: figures/main/pipeline_include.tex
\begin{figure*}[t]
    \centering
    \includegraphics[width=\textwidth]{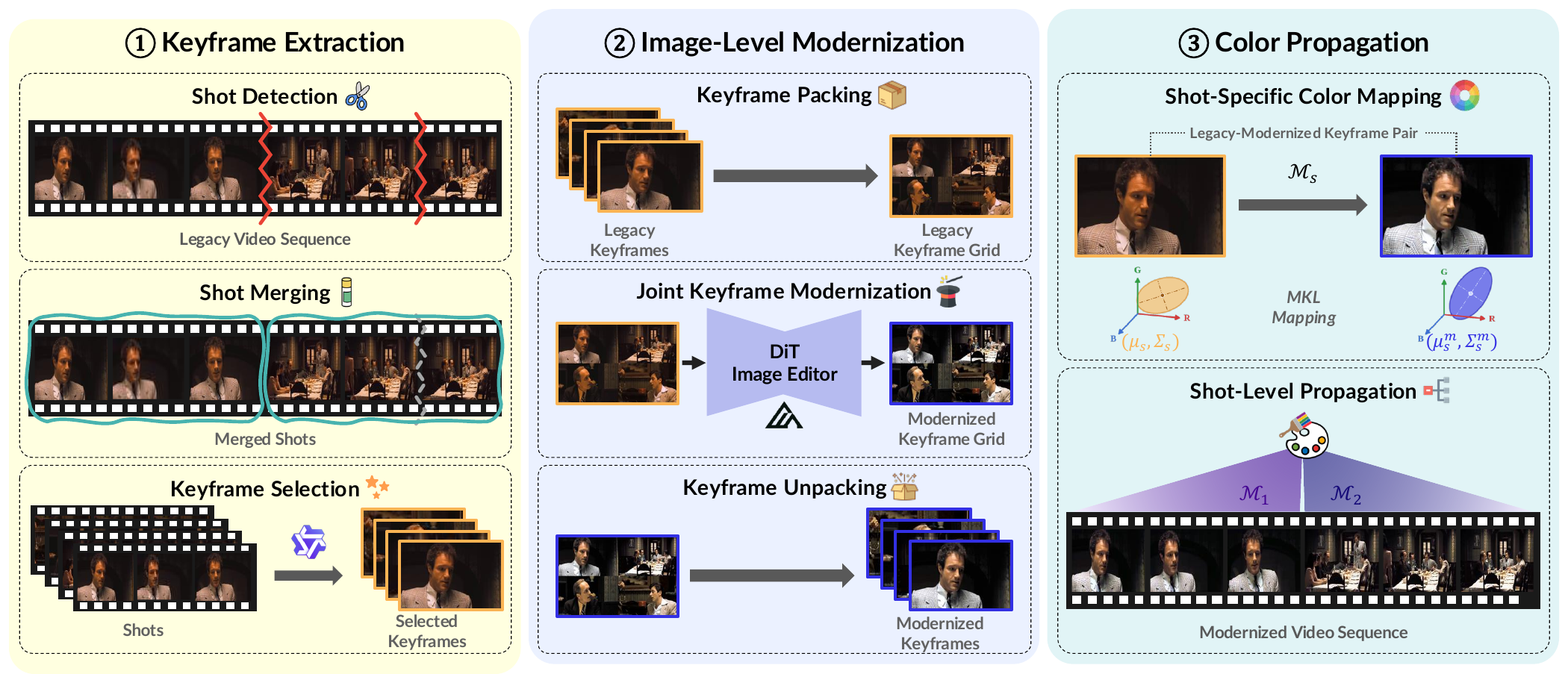}
    \caption{Overview of our \textbf{long video color modernization} framework. We first partition the input video into shots and select a representative keyframe from each shot. We then jointly modernize temporally adjacent keyframes. Finally, we estimate a shot-specific color mapping from each original--modernized keyframe pair and apply it to all frames in the corresponding shot.}
    \label{fig:framework}
\end{figure*}

%% file: figures/main/user_study_include.tex
\begin{figure}[t]
\centering
\includegraphics[width=\columnwidth]{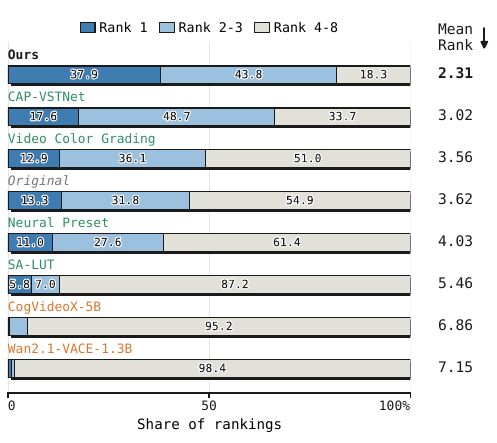}
\caption{Distribution of contemporary color preference rankings over 975 complete rankings. Each bar shows the proportions of Rank 1, Ranks 2--3, and Ranks 4--8, with the mean rank reported on the right.}
\label{fig:user_study_ranks}
\end{figure}

%% file: tables/main/quant.tex
\begin{table*}[t]
\centering
\small
\setlength{\tabcolsep}{4pt}
\begin{tabular}{@{}lccccccc@{}}
\toprule
Method
& Edge-SSIM $\uparrow$
& CDC $\downarrow$
& SW-IGC $\downarrow$
& DeQA $\uparrow$
& \revnum{Q-Align$_\mathrm{Q}$} $\uparrow$
& \revnum{Q-Align$_\mathrm{A}$} $\uparrow$
& \revfps{FPS} $\uparrow$ \\
\midrule
\refval{Original}
& \refval{0.966}
& \refval{0.56}
& \refval{0.19}
& \refval{2.73}
& \refval{3.16}
& \revnum{\refval{2.32}}
& \refval{--} \\
\midrule
\multicolumn{8}{@{}l}{\textcolor{grouporange}{\textit{End-to-End Video Editing}}} \\
CogVideoX-5B
& 0.439
& 0.77
& --
& 2.32
& 2.55
& \revnum{1.92}
& -- \\
Wan2.1-VACE-1.3B
& 0.535
& 1.37
& --
& 1.92
& 1.96
& \revnum{1.63}
& -- \\
\midrule
\multicolumn{8}{@{}l}{\textcolor{groupgreen}{\textit{Modular Color Propagation}}} \\
Neural Preset
& \textbf{0.886}
& \underline{0.56}
& 7.72
& 2.73
& \underline{3.11}
& \revnum{\textbf{2.34}}
& \revfps{\textbf{820}} \\
SA-LUT
& 0.745
& 0.77
& 7.23
& 2.49
& 2.91
& \revnum{2.23}
& \revfps{15} \\
Video Color Grading
& 0.843
& \textbf{0.54}
& \underline{6.45}
& 2.66
& 3.03
& \revnum{2.24}
& \revfps{23} \\
CAP-VSTNet
& 0.812
& 0.68
& 7.50
& \textbf{2.80}
& \textbf{3.13}
& \revnum{2.32}
& \revfps{82} \\
\midrule
\textbf{Ours}
& \underline{0.865}
& \underline{0.56}
& \textbf{5.90}
& \underline{2.78}
& \textbf{3.13}
& \revnum{\underline{2.33}}
& \revfps{\underline{816}} \\
\bottomrule
\end{tabular}
\caption{Complementary quantitative comparison on 100 videos from the Condensed Movies Dataset. The \revtext{unedited source video (Original)} is shown in gray for reference and excluded from ranking. CDC is reported in units of $10^{-2}$. SW-IGC is omitted for end-to-end methods since reliable source--output correspondence is unavailable. \revfps{FPS denotes the propagation throughput on an NVIDIA H200 GPU (see Appendix~\ref{sec:supp-runtime}).}}
\label{tab:quant}
\end{table*}

%% file: figures/main/qual_include.tex
\begin{figure*}[t]
\centering
\includegraphics[width=\textwidth]{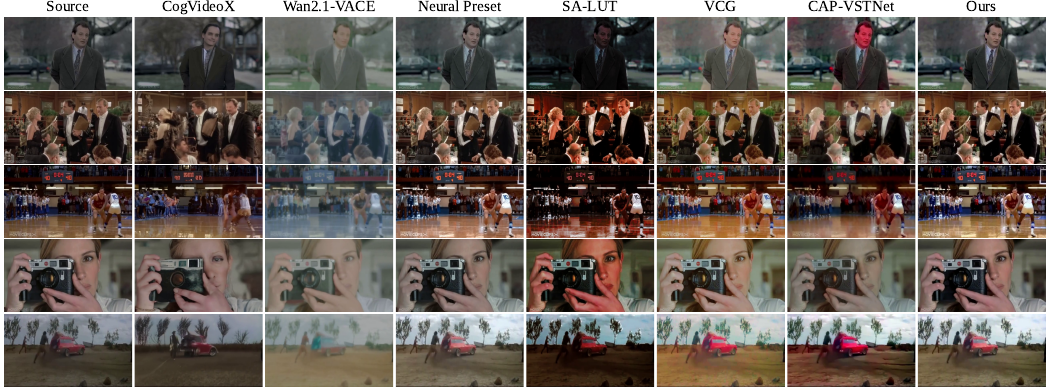}
\caption{Qualitative comparison of video modernization methods on representative frames.}
\label{fig:qual}
\end{figure*}

%% file: tables/main/ablation_1.tex
\begin{table}[t]
\centering
\small
\setlength{\tabcolsep}{4pt}
\begin{tabular}{@{}lcc@{}}
\toprule
Strategy & Invocations & SW-IGC $\downarrow$ \\
\midrule
Independent & $N$ & 7.23 \\
Independent + post-hoc smoothing
& $N$ & \textbf{4.46}$^{\dagger}$ \\
Packing (clustered)
& $\lceil N/4 \rceil$ & \underline{5.87} \\
Packing (consecutive)
& $\lceil N/4 \rceil$ & 5.90 \\
\bottomrule
\end{tabular}
\caption{Ablation of keyframe modernization strategies. $N$ denotes the number of shots in each video, and the packing strategies process up to four keyframes per model invocation. $^{\dagger}$Post-hoc smoothing directly regularizes the shot-specific mappings evaluated by SW-IGC.}
\label{tab:ablation_1}
\end{table}

%% file: figures/main/ablation_1_include.tex
\begin{figure*}[t]
\centering
\includegraphics[width=\textwidth]{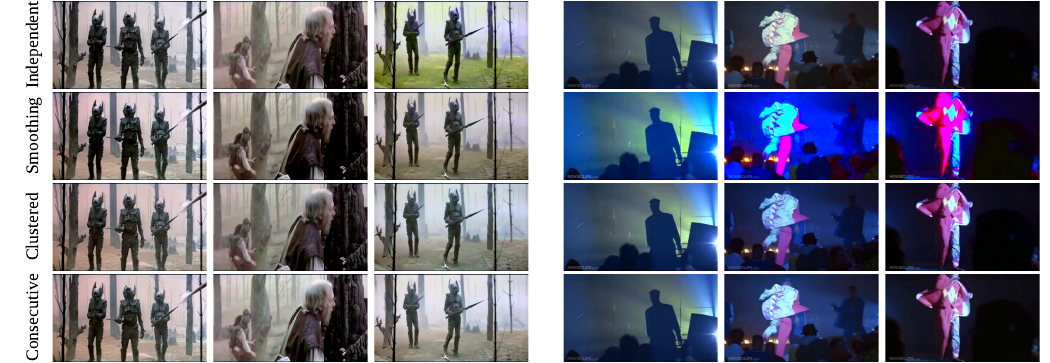}
\caption{Qualitative comparison of inter-shot consistency strategies for keyframe modernization. Independent modernization exhibits inter-shot color inconsistency, while post-hoc smoothing produces \revtext{unnatural} color distortions. The two joint strategies produce more coherent keyframe appearances across shots.}
\label{fig:ablation_1}
\end{figure*}

%% file: tables/main/ablation_2.tex
\begin{table}[t]
\centering
\small
\setlength{\tabcolsep}{8pt}
\begin{tabular}{lcc}
\toprule
Color mapping operator & SWD $\downarrow$ & Edge-SSIM $\uparrow$ \\
\midrule
Diagonal & 0.821 & \underline{0.884} \\
Histogram & \textbf{0.490} & 0.856 \\
MKL & \underline{0.742} & 0.883 \\
MKL + roll-off
& 0.748 & \textbf{0.885} \\
\bottomrule
\end{tabular}
\caption{Ablation of shot-specific color mapping operators. Each operator is fitted and evaluated on the same source keyframe and its modernized counterpart.}
\label{tab:ablation_2}
\end{table}

%% file: figures/main/ablation_2_include.tex
\begin{figure*}[t]
\centering
\includegraphics[width=\textwidth]{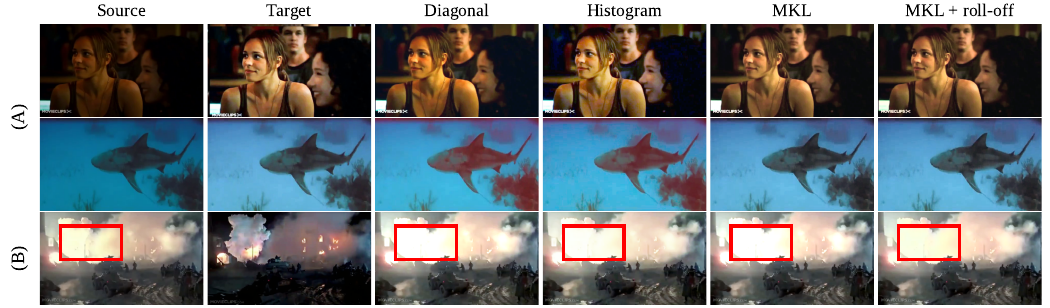}
\caption{Qualitative comparison of shot-specific color propagation operators. \textbf{(A)} Results obtained by applying each fitted mapping to its source keyframe. Top: \emph{Diagonal} fails to reproduce the target grade, whereas \emph{Histogram} introduces local color artifacts. Bottom: The per-channel mappings fail to reproduce the cross-channel target grade and produce strong color distortions, whereas \emph{MKL} closely matches it. \textbf{(B)} Results after propagating the fitted mapping to a brighter frame: hard clipping saturates highlights, while roll-off preserves their tonal variation.}
\label{fig:ablation_2}
\end{figure*}

%% file: supp.tex
\section{Overview}

This \revtext{appendix} provides additional details and analyses, organized as follows:

\begin{itemize}
    \item Implementation details of our framework in \revtext{Appendix}~\ref{sec:supp-implementation}.
    \item Detailed experimental settings in \revtext{Appendix}~\ref{sec:supp-settings}.
    \item Additional quantitative results in \revtext{Appendix}~\ref{sec:supp-quantitative}.
    \item Additional qualitative results in \revtext{Appendix}~\ref{sec:supp-qualitative}.
    \item Limitations and future directions in \revtext{Appendix}~\ref{sec:supp-limitations}.
\end{itemize}

\section{Implementation Details}
\label{sec:supp-implementation}

This section provides implementation details for the proposed framework, following the stage structure introduced in the main paper. Tab.~\ref{tab:supp-pipeline} summarizes the pipeline configurations used throughout all experiments.

\input{tables/supp/pipeline}

\subsection{Keyframe Extraction}

\paragraph{Shot detection.}
We adopt TransNet~V2~\revtext{\citep{transnetv2}} to detect candidate shot boundaries. TransNet~V2 is an efficient shot transition detection network that processes a sequence of neighboring frames and predicts a per-frame boundary probability, enabling fast shot segmentation of long videos. We threshold this probability at $0.1$, a deliberately recall-oriented choice: a missed boundary would cause a single color mapping to span distinct shots, whereas false positives merely over-segment the video and are corrected in the subsequent merging stage. Consecutive boundary predictions are consolidated into a single cut at the probability peak, yielding a contiguous, non-overlapping partition of the entire video.

\paragraph{Shot merging.}
The recall-oriented threshold may over-segment a continuous shot in the presence of flashes or rapid motion. To correct these false splits, we compare the frames immediately before and after each candidate boundary using the cosine similarity of their pooled DINOv2-S embeddings~\citep{dinov2}, and greedily merge adjacent segments whose similarity exceeds a merge threshold of $0.8$, so that a shot split into multiple fragments is fully recombined. This semantic criterion helps distinguish transient within-shot appearance changes from true boundaries between distinct shots.

\paragraph{Keyframe selection.}
The selected keyframe serves as both the modernization input and the color reference for its shot, so it must satisfy three requirements at once: it should be sharply focused, representative of the shot's content and color, and an actual film frame rather than a title card, poster, or other graphic that legacy clips frequently contain. These joint requirements are difficult to capture with a single hand-crafted criterion, so we cast selection as a holistic judgment by a vision--language model. From each shot, we uniformly sample 16 candidate frames and query Qwen3-VL~\citep{qwen3vl} once using the fixed instruction shown below; the returned index determines the shot's keyframe. Here, \texttt{\{N\}} is replaced by the number of candidate frames at query time.

\input{prompts/keyframe_selection}

\subsection{Image-Level Modernization}

\paragraph{Keyframe packing \& unpacking.}
We pack up to four temporally adjacent keyframes into a $2{\times}2$ image grid. When fewer than four keyframes are available, we fill the unused grid cells by repeating the available keyframes rather than using constant-valued padding. This avoids introducing large uniform regions that are atypical of natural images and may interfere with image editing. The packed grid is modernized in a single model invocation and subsequently unpacked into individual modernized keyframes, retaining only the outputs corresponding to the original keyframes.

\paragraph{Joint keyframe modernization.}
We use FLUX.1-Kontext~\revtext{\citep{fluxkontext}}, a \revtext{12B-parameter} flow-matching model for in-context image generation and editing, as the keyframe editor. Each packed grid is resized to the nearest supported resolution and edited in a single model invocation using the fixed instruction shown below, with a guidance scale of $2.5$ and $30$ sampling steps. Although the model is not specifically trained for grid-structured inputs, it generally applies coherent color and tonal adjustments across cells while preserving the content of each keyframe. The editor may slightly shift the grid boundaries, leading to minor cropping when the edited grid is unpacked. Such cropping has limited downstream impact because the propagation stage relies on global color statistics rather than pixelwise alignment.

\input{prompts/image_modernization}

\subsection{Color Propagation}

\paragraph{Shot-specific color mapping.}
For each shot, we fit the affine mapping in Eq.~\eqref{eq:mk_mapping} of the main paper using MKL~\citep{Monge_Kantorovitch} from the RGB distributions of the legacy and modernized keyframes. Means and covariances are computed over all pixels, with each covariance regularized as $\boldsymbol{\Sigma}\leftarrow \boldsymbol{\Sigma}+10^{-6}\mathbf{I}$. Since MKL matches global color distributions rather than pixel correspondences, fitting does not require spatial alignment between the two keyframes. The detailed formulation and closed-form solution are provided in the original work~\citep{Monge_Kantorovitch}.

\paragraph{Shot-level color propagation.}
The fitted affine mapping is applied uniformly to all frames in each shot. The mapping may produce RGB values above the valid range, for which direct clipping can flatten highlight detail and introduce hue shifts through uneven channel saturation. We therefore apply a luma-based highlight roll-off to each mapped pixel $\mathbf{x}'=(R',G',B')^\top$. Its Rec.~709 luma is computed from the gamma-encoded RGB values as
\begin{equation}
\label{eq:supp-luma}
Y' = 0.2126R' + 0.7152G' + 0.0722B'.
\end{equation}
Luma values above a fixed knee $k=0.75$ are smoothly compressed using
\begin{equation}
\label{eq:supp-rolloff}
\rho(Y') =
\begin{cases}
Y', & Y' \le k, \\[3pt]
k + (1-k)\tanh\!\left(\dfrac{Y'-k}{1-k}\right), & Y' > k.
\end{cases}
\end{equation}
For $Y'>k$, the RGB channels are uniformly rescaled by $\rho(Y')/Y'$, followed by clipping to the valid range. This shared scaling smoothly compresses highlight overshoots while preserving the relative channel ratios.

\subsection{Runtime}
\label{sec:supp-runtime}

We measure per-stage cost on ten videos from our dataset stratified by shot count, with one NVIDIA H200 GPU, and report each stage normalized by its natural cost unit in Tab.~\ref{tab:supp-runtime}; one-time model loading is excluded, as it amortizes over batch processing.

\input{tables/supp/runtime}

On this sample, end-to-end processing averages $10.9$ minutes per video, dominated by the per-shot VLM queries. Color propagation runs faster than real time even on the CPU\revfps{, measured as wall-clock throughput including video I/O}.

{\revfpson
\paragraph{Throughput protocol.}
Tab.~\ref{tab:quant} reports the propagation throughput alone, excluding the shared shot detection, keyframe selection, and keyframe modernization stages. All propagation methods are timed on the same GPU with a common protocol on 40 length-stratified shots ($10{,}334$ frames), with models preloaded, host--device transfers included, and video I/O excluded. All methods use a common batch of four frames, chosen so that SA-LUT fits in GPU memory, except Video Color Grading, which processes each shot as a whole. Ours and Neural Preset are within measurement uncertainty, with overlapping 95\% bootstrap intervals over shots. The end-to-end editors, timed on a representative clip, generate about $0.6$ frames per second.\par}

\section{Experiment Settings}
\label{sec:supp-settings}

\subsection{Dataset Curation}

We curate a subset of the Condensed Movies Dataset~\revtext{\citep{condensedmovies}} through a two-stage filtering process. First, we retain videos produced between the 1960s and the 2000s to capture diverse color characteristics across eras. Second, since our task focuses on modernizing existing color rather than colorizing black-and-white footage, we exclude videos consisting predominantly of black-and-white content. From the remaining videos, we randomly sample 20 per decade, resulting in an evaluation set of 100 videos.

\subsection{Baseline Details}

\paragraph{End-to-end video editing.}
For \revtext{Wan2.1-VACE-1.3B}~\citep{wanvace} and CogVideoX-5B~\citep{cogvideox}, we use the default inference configurations. We split each evaluation video into clips that match the frame length recommended for each model and concatenate the edited clips to reconstruct the full video. We use the same prompt for both video models:

\input{prompts/video_modernization}

\paragraph{Modular color propagation.}
For CAP-VSTNet~\citep{CAP_VSTNet}, SA-LUT~\citep{SA_LUT}, and Video Color Grading~\citep{Video_color_grading}, we use the default inference configurations. For Neural Preset~\citep{Neural_Preset}, as neither pretrained weights nor official code are publicly available, we reproduce the method based on the paper. Since the LUTs used to train Neural Preset were not publicly released, exact reproduction is not possible. We therefore assemble a separate LUT collection for our implementation.

\subsection{User Study Details}

We recruit 65 participants through the Prolific website for human evaluation. Fig.~\ref{fig:instruction} presents the instructions provided to the participants. We prepare 15 video sets, each displayed as a grid of eight videos playing simultaneously\revtext{,} comprising the original video and the outputs of the evaluated methods. We randomize both the order of the video sets and the positions of all videos within each grid, including the original. After viewing each grid, as illustrated in Fig.~\ref{fig:ranking}, participants rank the videos according to \revtext{two criteria: modern color appearance, judged on color alone, and temporal consistency of colors and content}.

\input{tables/supp/user_study}

\section{Additional Quantitative Results}
\label{sec:supp-quantitative}

\subsection{VLMs Cannot Reliably Assess Modern Color}

Our primary evaluation is a modern color preference study. Determining whether the dated colors of legacy footage have been sufficiently transformed into a contemporary appearance is inherently subjective, suggesting two possible evaluation strategies: VLM-based evaluation and human evaluation. However, we find that VLM-based judgments are inconsistent and therefore unreliable. Based on Fig.~\ref{fig:vlm-score-matrix} and Tab.~\ref{tab:vlm_order_stability}, we exclude VLM-based evaluation from our evaluation and instead recruit human participants, highlighting that assessing modern color appearance requires direct human judgment rather than relying solely on existing automated evaluators.

\input{tables/supp/vlm_order_stability}

We assess the reliability of VLM-based evaluation using two commonly adopted protocols: direct scoring and pairwise preference comparison under a two-alternative forced-choice (2AFC) setting. We use Qwen3-VL~\citep{qwen3vl} as the judge and evaluate it on 15 video sets sampled from the evaluation set. Each set contains the source video and the outputs of all compared methods, from which we uniformly sample 24 frames per video.

For direct scoring, we divide the 24 sampled frames for each method into six chunks of four frames and ask the VLM to assign a modern color score from 1 to 5 to each chunk. For video \(v\), method \(m\), and chunk \(c\), let \(s_{v,m,c}\) be the VLM score. Each method receives six chunk scores, so its video-level score is:
\[
s_{v,m}
=
\frac{1}{6}\sum_{c=1}^{6}s_{v,m,c}.
\]

The average pairwise-difference matrix of these 15 videos is defined as follows:
\[
\bar D(i,j)
=
\frac{1}{15}\sum_{v=1}^{15}D_v(i,j)
=
\frac{1}{15}\sum_{v=1}^{15}
\left|s_{v,m_i}-s_{v,m_j}\right|.
\]

\revtext{Across 15 videos, the eight method means occupy only the interval 2.2889--2.8889 on the 1--5 scale, and Fig.~\ref{fig:vlm-score-matrix} shows a mean pairwise absolute difference of only 0.3544.} The result shows that many visually different color appearances receive nearly equivalent scores, indicating that \revtext{the} VLM does not provide sufficiently discriminative evidence for comparing color modernization quality.

\input{figures/supp/score_matrix_include}

Tab.~\ref{tab:vlm_order_stability} shows results in \revtext{the 2AFC setting}. For each method pair, we evaluate temporally aligned frames in both A/B orders and consider the result consistent only if the same underlying method wins. Swapping the order changes the winner for 158 of 420 pairs (37.6\%). Candidate B is selected 65\% of the time, 30 percentage points more often than A, revealing a substantial preference for the second position. \revtext{The} VLM sometimes changes its preference when the positions of the two candidates are swapped, revealing substantial positional inconsistency.
\input{prompts/vlm_scoring}
\input{prompts/vlm_2afc}

\subsection{User Study Results}

We conduct a user study to evaluate whether each method successfully transforms the colors of legacy footage into a modern style. As shown in Fig.~\ref{fig:vlm-score-matrix} and Tab.~\ref{tab:vlm_order_stability}, we find that VLM-based evaluation does not provide sufficiently consistent judgments. Determining whether a method produces modern colors and improves upon the original footage is inherently subjective.

We collect a total of 975 rankings from 65 participants and evaluate the results using Mean Rank, Rank-1, and Top-3 rates. Tab.~\ref{tab:user_study} shows that our method outperforms all baselines across all three measures, indicating that it most successfully transforms the dated colors of legacy footage into a contemporary appearance. Interestingly, Neural Preset, SA-LUT, CogVideoX-5B, and Wan2.1-VACE-1.3B perform worse than the unedited original video. This suggests that these methods either fail to produce a sufficiently contemporary color appearance or introduce content instability that reduces user preference.

\section{Additional Qualitative Results}
\label{sec:supp-qualitative}

\subsection{Comparison of Keyframe Selection Strategies}

Selecting an appropriate keyframe is critical, as it serves both as the input to the image editor and as the color reference for the entire shot. We compare our vision--language model-based selector with four alternatives based on heuristic, statistical, and learned criteria. \emph{First frame} selects the first frame of each shot, while the \revtext{\emph{color medoid}} selects the frame whose mean color is closest to the shot-level average. \emph{MUSIQ}~\citep{musiq} selects the frame with the highest no-reference image-quality score, and the \emph{DINOv2 medoid}~\citep{dinov2} selects the frame closest to the mean DINOv2 representation of the shot.

\input{figures/supp/keyframe_selection_include}

Fig.~\ref{fig:supp-kfsel} illustrates representative limitations of these selection criteria, with failure cases outlined in red. First frame is sensitive to shot transitions and rapid motion, and may select blurred frames or views captured before the main subject appears. MUSIQ prioritizes perceptual quality but does not assess shot representativeness, occasionally favoring title cards or poster-like graphics over actual scene content. The DINOv2 medoid favors semantically representative frames but does not explicitly account for image quality, while the color medoid similarly captures color centrality without assessing visual quality or whether the frame contains informative scene content. In contrast, our vision--language model jointly considers visual quality, shot representativeness, and whether each candidate is a valid content frame.

\subsection{Comparison with Baselines}

Fig.~\ref{fig:supp-qual-1} and Fig.~\ref{fig:supp-qual-2} present additional qualitative comparisons between our method and the evaluated baselines on frames sampled from the evaluation set, covering diverse content and visual conditions.

\section{Limitations and Future Directions}
\label{sec:supp-limitations}

Our framework has three main limitations. First, joint modernization is restricted to groups of up to four temporally adjacent keyframes. It therefore promotes local cross-shot consistency but does not explicitly coordinate the appearance of distant or recurring shots throughout a long video. Second, shot-level propagation relies on a single global affine color mapping. While this design efficiently preserves spatial structure and temporal stability, it cannot reproduce spatially varying edits or nonlinear local adjustments introduced by the image editor. Third, the method cannot reliably distinguish unintended color degradation from deliberate stylistic choices, such as muted palettes or scene-specific color casts, and may therefore alter the intended visual style.

Future work could explore hierarchical or memory-based coordination across keyframes to improve long-range coherence while preserving shot-specific flexibility. Spatially adaptive propagation may better capture local edits without sacrificing temporal consistency. Incorporating user input or reference images could also help preserve the intended visual style by specifying the desired degree and direction of modernization. Broader evaluation across genres, production formats, and viewer populations would further clarify how perceptions of contemporary color vary across viewing contexts.

\input{figures/supp/user_study_interface}

\input{figures/supp/qual_supp_1_include}
\input{figures/supp/qual_supp_2_include}

%% file: tables/supp/pipeline.tex
\begin{table}[t]
\centering
\small
\setlength{\tabcolsep}{6pt}
\begin{tabular}{@{}ll@{}}
\toprule
\textbf{Configuration} & \textbf{Value} \\
\midrule
\multicolumn{2}{@{}l}{\textcolor{royalblue}{\emph{Keyframe Extraction}}} \\
\addlinespace[2pt]
\multicolumn{2}{@{}l}{\quad \textbf{Shot detection}} \\
\quad\enspace detector           & TransNet~V2 \\
\quad\enspace boundary threshold & 0.1 \\
\addlinespace[5pt]
\multicolumn{2}{@{}l}{\quad \textbf{Shot merging}} \\
\quad\enspace boundary encoder   & DINOv2-S \\
\quad\enspace merge threshold    & 0.8 \\
\addlinespace[5pt]
\multicolumn{2}{@{}l}{\quad \textbf{Keyframe selection}} \\
\quad\enspace selector           & Qwen3-VL-8B-Instruct \\
\quad\enspace candidates per shot & 16 \\
\midrule
\multicolumn{2}{@{}l}{\textcolor{royalblue}{\emph{Image-Level Modernization}}} \\
\addlinespace[2pt]
\multicolumn{2}{@{}l}{\quad \textbf{Keyframe packing \& unpacking}} \\
\quad\enspace strategy           & $2{\times}2$, consecutive \\
\quad\enspace padding            & replicate \\
\addlinespace[5pt]
\multicolumn{2}{@{}l}{\quad \textbf{Joint keyframe modernization}} \\
\quad\enspace editor             & FLUX.1-Kontext-dev \\
\quad\enspace guidance scale     & 2.5 \\
\quad\enspace inference steps     & 30 \\
\midrule
\multicolumn{2}{@{}l}{\textcolor{royalblue}{\emph{Color Propagation}}} \\
\addlinespace[2pt]
\multicolumn{2}{@{}l}{\quad \textbf{Shot-specific color mapping}} \\
\quad\enspace mapping operator   & MKL (RGB) \\
\addlinespace[5pt]
\multicolumn{2}{@{}l}{\quad \textbf{Shot-level color propagation}} \\
\quad\enspace roll-off knee $k$  & 0.75 \\
\bottomrule
\end{tabular}
\caption{Pipeline configurations used for all experiments.}
\label{tab:supp-pipeline}
\end{table}

%% file: prompts/keyframe_selection.tex
\begin{center}
\centering
\fcolorbox{black!45}{black!3}{%
\begin{minipage}{\dimexpr\columnwidth-2\fboxsep-2\fboxrule\relax}
\textbf{Keyframe Selection Prompt}

{\small\ttfamily\raggedright
\textcolor{royalblue}{["These are \{N\} frames sampled in order from one continuous shot of an old movie, with indices ranging from 0 to \revtext{\{N-1\}}. Pick the ONE best frame to modernize and use as the color reference for the whole shot: (1) sharply in focus, NO motion blur; (2) clearly shows the main subject, representative of the shot; (3) a real film frame, NOT an end-card / movie poster / thumbnail grid / logo / text graphic. Answer with ONLY the index number (0--\{N-1\}), nothing else."]}\par}
\end{minipage}}
\end{center}

%% file: prompts/image_modernization.tex
\begin{center}
\centering
\fcolorbox{black!45}{black!3}{%
\begin{minipage}{\dimexpr\columnwidth-2\fboxsep-2\fboxrule\relax}
\textbf{Joint Keyframe Modernization Prompt}

{\small\ttfamily\raggedright
\textcolor{royalblue}{["Modernize the color of each old film frame in this image grid to a natural, contemporary look. Apply clean, balanced colors and natural skin tones consistently across all grid cells, removing dated or faded color casts. Preserve the content, identity, composition, framing, spatial layout, and original lighting and time of day of every frame. Keep the grid structure and cell boundaries unchanged. Do not add, remove, or hallucinate any content."]}\par}
\end{minipage}}
\end{center}

%% file: tables/supp/runtime.tex
\begin{table}[t]
\centering
\small
\setlength{\tabcolsep}{6pt}
\begin{tabular}{@{}lll@{}}
\toprule
\textbf{Stage} & \textbf{Device} & \textbf{Cost} \\
\midrule
Shot detection            & CPU & $\sim$24\,s / min of video \\
Shot merging              & GPU & $\sim$0.9\,s / shot \\
Keyframe selection        & GPU & $\sim$13\,s / shot \\
Image-level modernization & GPU & $\sim$14\,s / grid \\
Color propagation         & CPU & $\sim$80 FPS \\
\bottomrule
\end{tabular}
\caption{Per-stage processing cost, pooled over ten videos from our dataset.}
\label{tab:supp-runtime}
\end{table}

%% file: prompts/video_modernization.tex
\begin{center}
\centering
\fcolorbox{black!45}{black!3}{%
\begin{minipage}{\dimexpr\columnwidth-2\fboxsep-2\fboxrule\relax}
\textbf{Modernization Prompt for Video Models}

{\small\ttfamily\raggedright
\textcolor{royalblue}{["Modernize the color of this old film to a natural, contemporary look: clean, balanced, natural colors and skin tones, without dated or faded color casts. Keep the content, identity, framing, and the original lighting and time of day unchanged; do not add, remove, or hallucinate anything."]}\par}

\vspace{\baselineskip}

\textbf{Negative Prompt for Video Models}

{\small\ttfamily\raggedright
\textcolor{royalblue}{["overexposed, oversaturated, blurry, low quality, jpeg artifacts, warped face, changed identity, changed pose, changed clothing, changed framing, hallucinated objects, extra limbs, deformed hands, subtitles, watermark, text"]}\par}
\end{minipage}}
\end{center}

%% file: tables/supp/user_study.tex
\begin{table}[t]
\centering
\small
\begin{tabular}{@{}lccc@{}}
\toprule
\begin{tabular}[c]{@{}l@{}}Method\end{tabular} &
\begin{tabular}[c]{@{}c@{}}Mean\\Rank $\downarrow$\end{tabular} &
\begin{tabular}[c]{@{}c@{}}Rank-1 $\uparrow$\end{tabular} &
\begin{tabular}[c]{@{}c@{}}Top-3 $\uparrow$\end{tabular} \\
\midrule
\textbf{Ours} & \textbf{2.313} & \textbf{37.9\%} & \textbf{81.7\%} \\
CAP-VSTNet & \underline{3.015} & \underline{17.6\%} & \underline{66.3\%} \\
Video Color Grading & 3.559 & 12.9\% & 49.0\% \\
Original & 3.619 & 13.3\% & 45.1\% \\
Neural Preset & 4.028 & 11.0\% & 38.6\% \\
SA-LUT & 5.461 & 5.8\% & 12.8\% \\
CogVideoX-5B & 6.859 & 0.4\% & 4.8\% \\
Wan2.1-VACE-1.3B & 7.146 & 0.9\% & 1.6\% \\
\bottomrule
\end{tabular}
\caption{Human evaluation of modern color preference over 975 complete rankings (65 participants $\times$ 15 videos). Mean Rank uses all placements from 1 to 8; Rank-1 and Top-3 denote the percentage of rankings in the corresponding range.}
\label{tab:user_study}
\end{table}

%% file: tables/supp/vlm_order_stability.tex
\begin{table}[t]
\centering
\resizebox{\columnwidth}{!}{%
\begin{tabular}{l r}
\toprule
Metric & Value \\
\midrule
Videos & 15 \\
Methods per video & 8 \\
Unordered method pairs per video ($\binom{8}{2}$) & 28 \\
2AFC queries per pair (forward + reversed order) & 2 \\
Total VLM queries & 840 \\
Complete swapped-order comparisons & 420 \\
Order-consistent comparisons & 262 / 420 (62.4\%) \\
Order-inconsistent comparisons & 158 / 420 (37.6\%) \\
Candidate A selection rate & 35.0\% \\
Candidate B selection rate & 65.0\% \\
\bottomrule
\end{tabular}
}
\caption{Order-swap stability of the VLM-based two-alternative forced-choice
(2AFC) color judge. For each unordered method pair, we query the judge twice: once in the original A/B order and once with the two candidates swapped. A comparison is consistent only when both queries select the same underlying method after mapping the A/B labels back to method identities.}
\label{tab:vlm_order_stability}
\end{table}

%% file: figures/supp/score_matrix_include.tex
\begin{figure}[t]
\centering
\includegraphics[width=\columnwidth]{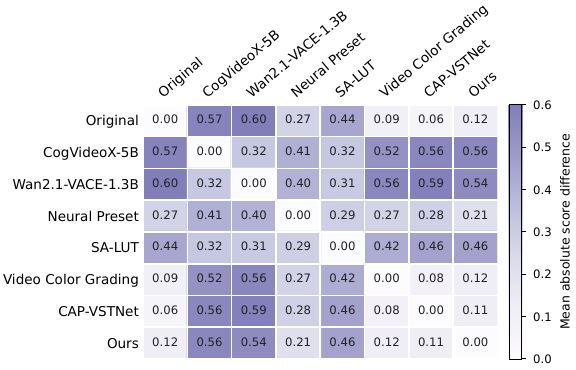}
\caption{\revtext{Average pairwise absolute differences between the VLM scores of the eight evaluated methods across 15 videos.}}
\label{fig:vlm-score-matrix}
\end{figure}

%% file: prompts/vlm_scoring.tex
\definecolor{promptborder}{RGB}{214,170,54}
\definecolor{promptbackground}{RGB}{255,249,218}
\definecolor{prompttitle}{RGB}{250,224,140}
\definecolor{prompttext}{RGB}{74,57,24}

\begin{figure*}[t]
\centering
\begingroup
\setlength{\fboxsep}{4pt}%
\noindent\fcolorbox{promptborder}{prompttitle}{%
\parbox{\dimexpr\linewidth-2\fboxsep-2\fboxrule\relax}{%
\centering\bfseries VLM-based Modern Color Scoring Prompt}}
\begin{lstlisting}[
  basicstyle=\scriptsize\ttfamily\color{prompttext},
  backgroundcolor=\color{promptbackground},
  frame=single,
  rulecolor=\color{promptborder},
  framerule=0.6pt,
  framesep=7pt,
  aboveskip=0pt,
  belowskip=0pt,
  breaklines=true,
  breakatwhitespace=false,
  columns=fullflexible,
  keepspaces=true,
  showstringspaces=false,
  upquote=true,
  escapeinside={(*@}{@*)}
]
(*@\textbf{System Prompt}@*)
["You are an expert evaluator for a video color-modernization study.
Your task is to evaluate ONLY the color treatment, not the content or image quality.
Use the full (*@\revtext{1-5}@*) scale. Do not be overly harsh or overly generous.
A score of 3 means acceptable or mixed color modernization, not failure.

Do NOT evaluate:
- whether the subject, clothes, objects, props, architecture, or location look old or modern
- image resolution
- sharpness or blur
- noise
- compression artifacts
- object content
- scene composition
- acting, story, aesthetics, or cinematic framing
- non-color effects
- temporal consistency or motion continuity"]

(*@\textbf{Prompt}@*)
["You will receive sampled frames from one generated video.
Evaluate whether the color treatment looks modernized, natural, and color-balanced.
Focus only on color appearance.

Evaluate ONLY:

- natural_saturation: colors are neither artificially oversaturated nor unnecessarily desaturated.
- cast_neutrality: no unintended yellow, green, turquoise, blue or magenta cast of the old video and has a neutral color tone.
- naturalness: semantically familiar colors such as daytime, skin, objects, and neutrals look plausible

Overall Scoring:
Use a 1-5 scale for overall_color_modernity.

5 = color looks clean and acceptable for contemporary viewing
4 = mostly acceptable, with minor old-color artifacts
3 = mixed; some old-color artifacts are clearly visible
2 = mostly outdated; old-color artifacts dominate
1 = severely outdated or color-biased

- A clean but ordinary color treatment can receive 4 or 5.
- Use low acceptability scores only when outdated colors are clearly visible.

Score overall_color_modernity from 1 to 5.

Example output:
{
  "frame_ids": [the exact frame_ids_in_order list],
  "overall_color_modernity": 3,
  "reason": "Although the color treatment appears somewhat modern, it is not fully natural."
}"]
\end{lstlisting}
\endgroup
\end{figure*}

%% file: prompts/vlm_2afc.tex
\definecolor{promptborder}{RGB}{139,181,105}
\definecolor{promptbackground}{RGB}{242,250,229}
\definecolor{prompttitle}{RGB}{211,235,174}
\definecolor{prompttext}{RGB}{48,76,38}

\begin{figure*}[t]
\centering
\begingroup
\setlength{\fboxsep}{4pt}%
\noindent\fcolorbox{promptborder}{prompttitle}{%
\parbox{\dimexpr\linewidth-2\fboxsep-2\fboxrule\relax}{%
\centering\bfseries VLM-based Pairwise Preference Comparison Prompt}}
\begin{lstlisting}[
  basicstyle=\scriptsize\ttfamily\color{prompttext},
  backgroundcolor=\color{promptbackground},
  frame=single,
  rulecolor=\color{promptborder},
  framerule=0.6pt,
  framesep=7pt,
  aboveskip=0pt,
  belowskip=0pt,
  breaklines=true,
  breakatwhitespace=false,
  columns=fullflexible,
  keepspaces=true,
  showstringspaces=false,
  upquote=true,
  escapeinside={(*@}{@*)}
]
(*@\textbf{System Prompt}@*)
["You are an expert evaluator for a video color-modernization study.
You will be given two matching frames from two candidate videos.
Your task is to evaluate ONLY the color treatment, not the content or image quality.
Use the full distinction between candidates. You must choose exactly one winner.

Do NOT evaluate:
- whether the subject, clothes, objects, props, architecture, or location look old or modern
- image resolution
- sharpness or blur
- noise
- compression artifacts
- object content
- scene composition
- acting, story, aesthetics, or cinematic framing
- non-color effects
- temporal consistency or motion continuity"]

(*@\textbf{Prompt}@*)
["Compare Candidate A and Candidate B for color modernization.
Focus only on color appearance.

Evaluate ONLY:
- natural_saturation: colors are neither artificially oversaturated nor unnecessarily desaturated.
- cast_neutrality: no unintended yellow, green, turquoise, blue or magenta cast of the old video and has a neutral color
tone.
- naturalness: semantically familiar colors such as daytime, skin, objects, and neutrals look plausible

Choose exactly one winner: A or B. Do not return TIE.
If the candidates are very close, choose the one with slightly more plausible, neutral, and modern color treatment.

Example output:
{
  "winner": "A",
  "reason": "Candidate A has more neutral and natural color balance, with fewer outdated casts than Candidate B."
}"]
\end{lstlisting}
\endgroup
\end{figure*}

%% file: figures/supp/keyframe_selection_include.tex
\begin{figure*}[t]
\centering
\includegraphics[width=\textwidth]{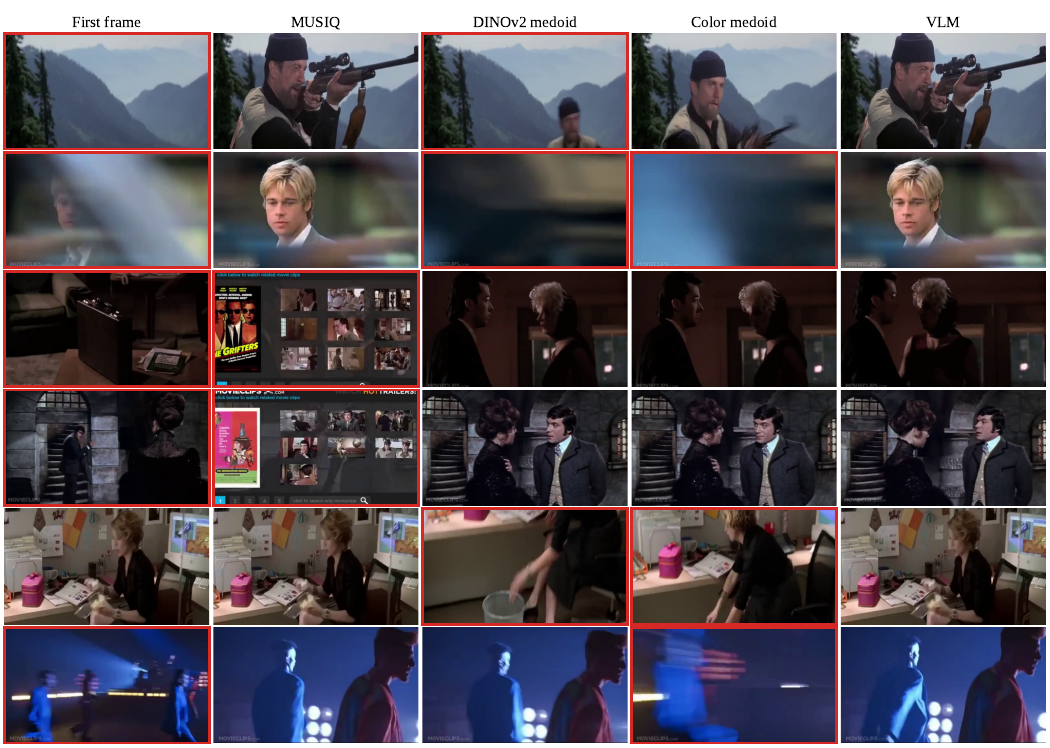}
\caption{Comparison of keyframe selection methods. Each row shows the keyframes selected by different methods for the same shot; failure cases are outlined in red.}
\label{fig:supp-kfsel}
\end{figure*}

%% file: figures/supp/user_study_interface.tex
\begin{figure*}[p]
\centering
\includegraphics[width=0.72\textwidth]{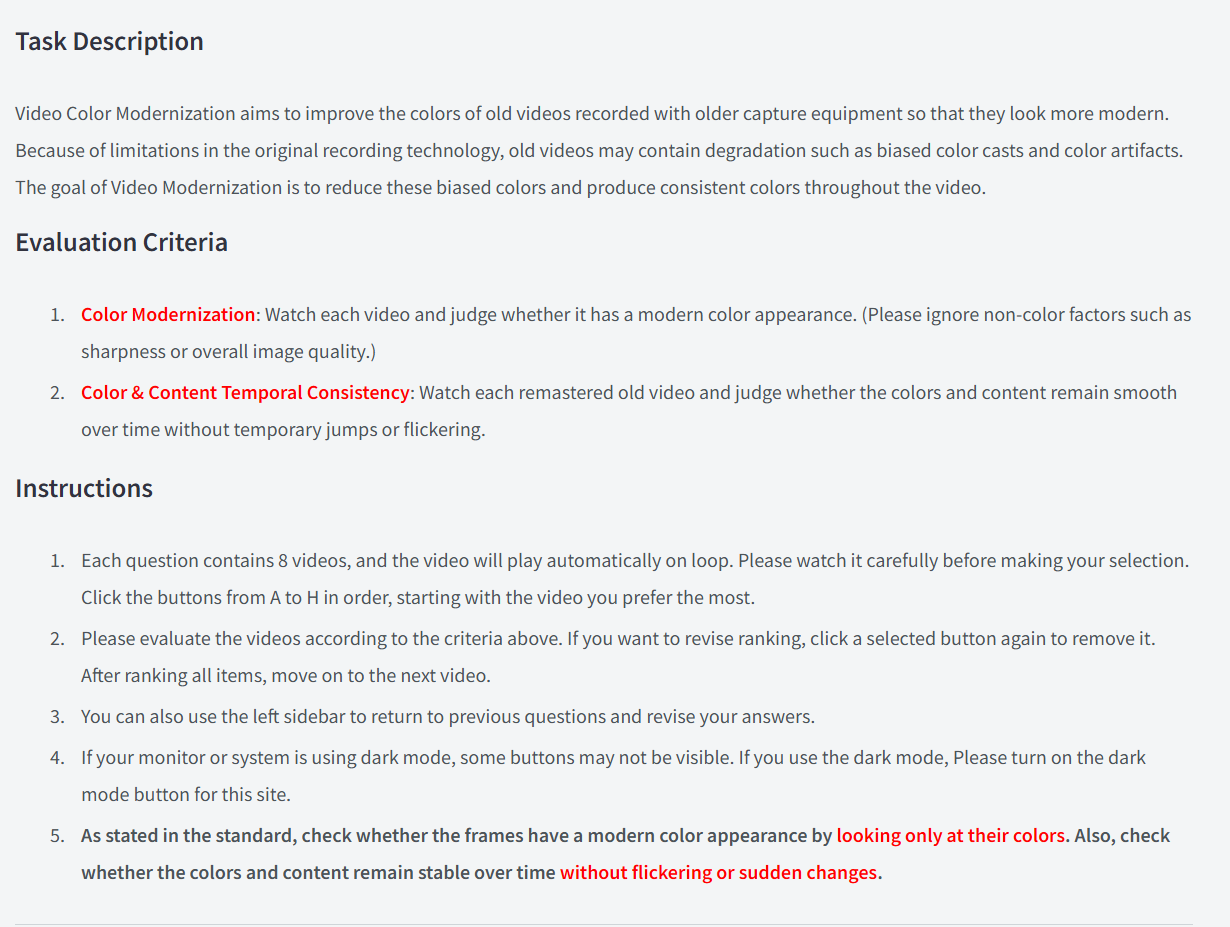}
\caption{Instructions presented to participants in the user study.}
\label{fig:instruction}

\vspace{0.5em}

\includegraphics[width=0.72\textwidth]{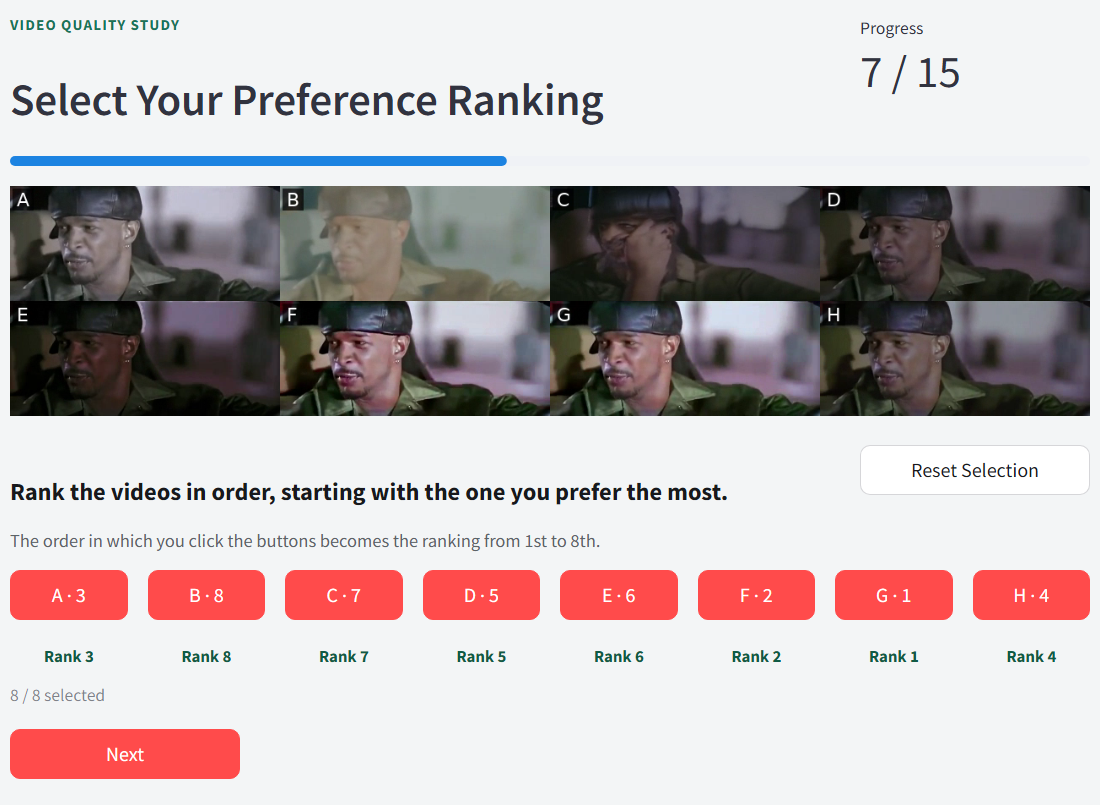}
\caption{Interface used by participants to rank the eight video versions.}
\label{fig:ranking}
\end{figure*}

%% file: figures/supp/qual_supp_1_include.tex
\begin{figure*}[t]
\centering
\includegraphics[width=\textwidth]{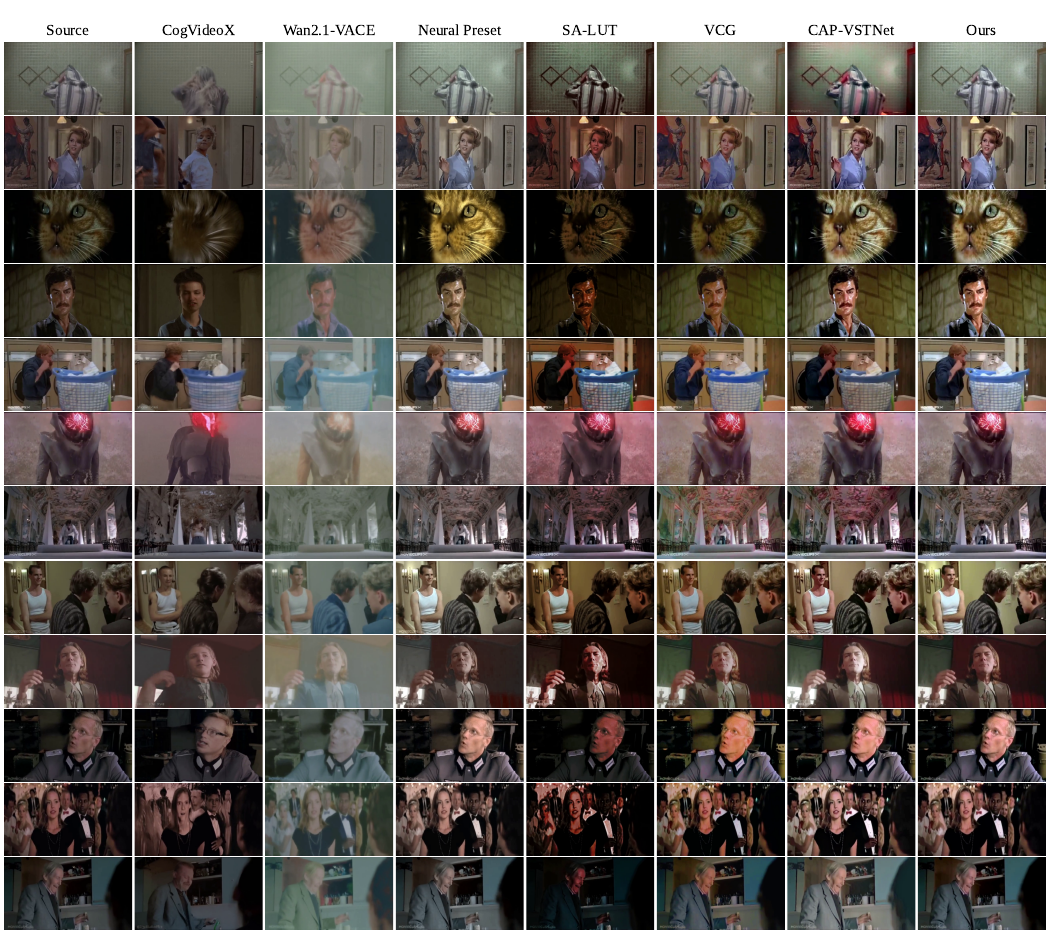}
\caption{Additional qualitative comparisons between our method and the evaluated baselines on diverse frames from the evaluation set (Part 1).}
\label{fig:supp-qual-1}
\end{figure*}

%% file: figures/supp/qual_supp_2_include.tex
\begin{figure*}[t]
\centering
\includegraphics[width=\textwidth]{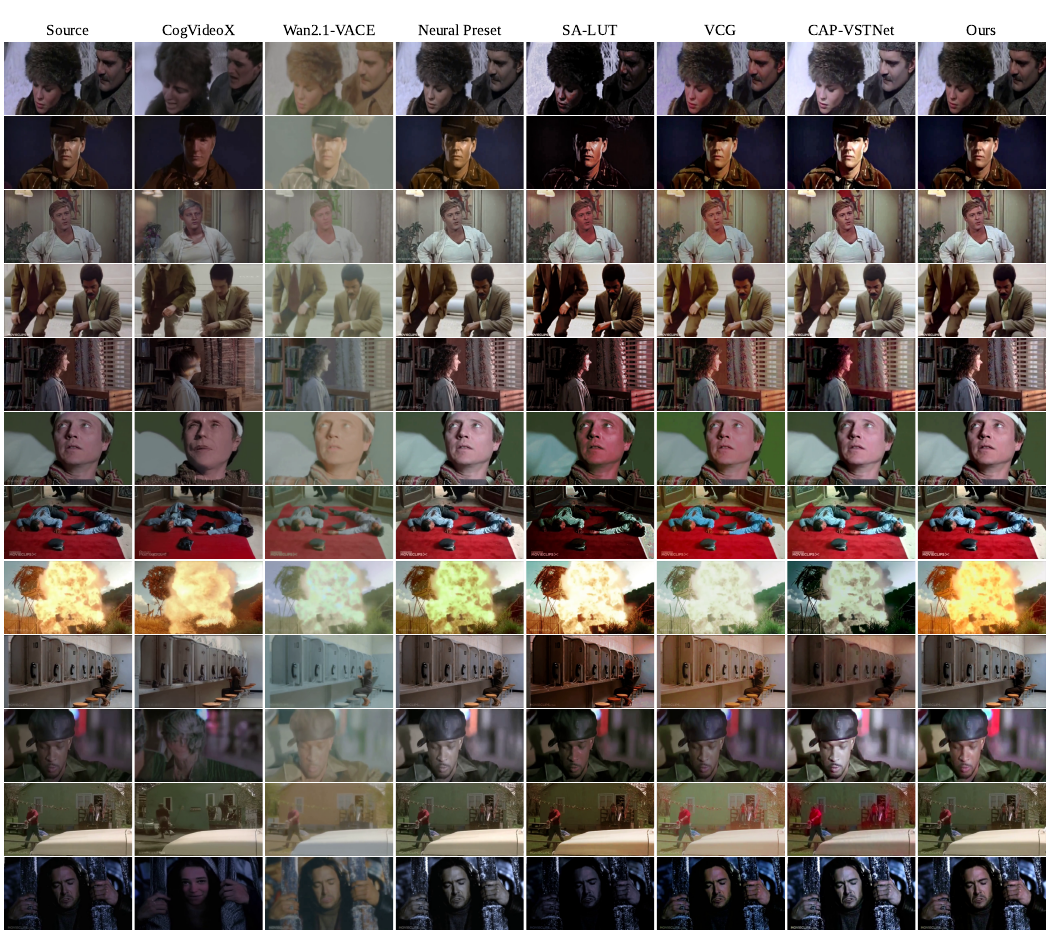}
\caption{Additional qualitative comparisons between our method and the evaluated baselines on diverse frames from the evaluation set (Part 2).}
\label{fig:supp-qual-2}
\end{figure*}